%% file: iccv19.tex
\documentclass[10pt,twocolumn,letterpaper]{article}

\usepackage{iccv}
\usepackage{times}
\usepackage{epsfig}
\usepackage{graphicx}
\usepackage{amsmath}
\usepackage{amssymb}

\usepackage{multirow}
\usepackage{comment}
\usepackage{bm} % bold math
\usepackage[table]{xcolor}
\usepackage{subfig}

\newcounter{ablStdModel} % model indices in ablation study
\newcommand{\ablStdModel}{\refstepcounter{ablStdModel}\theablStdModel}
\newcommand{\keypoint}[1]{\vspace{.1cm}\noindent\textbf{#1}}
\newcommand{\red}[1]{\textcolor{red}{#1}}
\newcommand{\blue}[1]{\textcolor{blue}{#1}}
\newcommand{\green}[1]{\textcolor{green}{#1}}
\newcommand{\std}[1]{\scriptsize$\pm${#1}}
\usepackage[pagebackref=true,breaklinks=true,letterpaper=true,colorlinks,bookmarks=false]{hyperref}

\iccvfinalcopy % *** Uncomment this line for the final submission

\def\iccvPaperID{2964} % *** Enter the ICCV Paper ID here

\ificcvfinal\fi
\begin{document}

%%%%%%%%% TITLE
\title{Omni-Scale Feature Learning for Person Re-Identification}

\author{
Kaiyang Zhou$^{1}$ \quad
Yongxin Yang$^{1}$ \quad
Andrea Cavallaro$^{2}$ \quad
Tao Xiang$^{1,3}$
\vspace{0.5em} \\
$^1$University of Surrey \quad
$^2$Queen Mary University of London \\
$^3$Samsung AI Center, Cambridge \\
{\tt\small
\{k.zhou, yongxin.yang, t.xiang\}@surrey.ac.uk \quad
a.cavallaro@qmul.ac.uk
}
}

\maketitle
%\thispagestyle{empty}

\input{main.tex}
\input{supp.tex}

{\small
\bibliographystyle{ieee}
\bibliography{reference}
}

\end{document}

%% file: main.tex
%%%%%%%%% ABSTRACT
\begin{abstract}
As an instance-level recognition problem, person re-identification (re-ID) relies on discriminative features, which not only capture different spatial scales but also encapsulate an arbitrary combination of multiple scales. We call features of both homogeneous and heterogeneous scales omni-scale features. In this paper, a novel deep re-ID CNN is designed, termed omni-scale network (OSNet), for omni-scale feature learning. This is achieved by designing a residual block composed of multiple convolutional streams, each detecting features at a certain scale. Importantly, a novel unified aggregation gate is introduced to dynamically fuse multi-scale features with input-dependent channel-wise weights. To efficiently learn spatial-channel correlations and avoid overfitting, the building block uses pointwise and depthwise convolutions. By stacking such block layer-by-layer, our OSNet is extremely lightweight and can be trained from scratch on existing re-ID benchmarks. Despite its small model size, OSNet achieves state-of-the-art performance on six person re-ID datasets, outperforming most large-sized models, often by a clear margin. Code and models are available at: \url{https://github.com/KaiyangZhou/deep-person-reid}.
\end{abstract}

%%%%%%%%% BODY TEXT
\section{Introduction} \label{sec:intro}
\begin{figure}[t!]
\centering
\includegraphics[width=0.8\columnwidth]{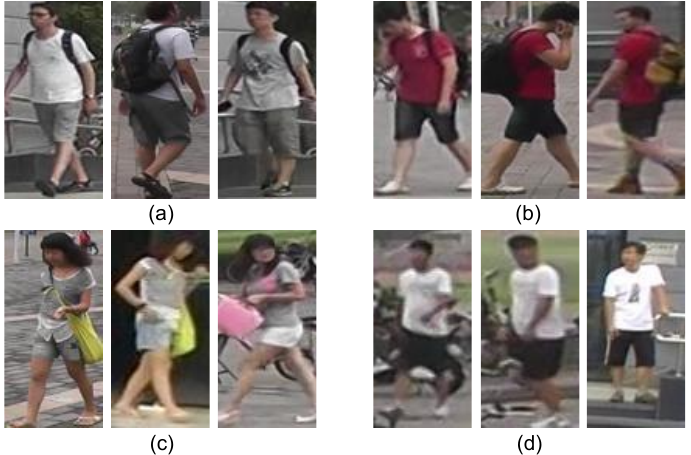}
\caption{Person re-ID is a hard problem, as exemplified by the four triplets of images above. Each sub-figure shows, from left to right, the query image, a true match and an impostor/false match.}
\vspace{-0.2cm}
\label{fig:examples}
\end{figure}

Person re-identification (re-ID), a fundamental task in distributed multi-camera surveillance, aims to match people appearing in different non-overlapping camera views. As an instance-level recognition problem, person re-ID faces two major challenges as illustrated in Fig.~\ref{fig:examples}. First, the intra-class (instance/identity) variations are typically big due to the changes of camera viewing conditions. For instance, both people in Figs.~\ref{fig:examples}(a) and (b) carry a backpack; the view change across cameras (frontal to back) brings large appearance changes in the backpack area, making matching the same person difficult. Second, there are also small inter-class variations -- people in public space often wear similar clothes; from a distance as typically in surveillance videos, they can look incredibly similar (see the impostors for all four people in Fig.~\ref{fig:examples}).

To overcome these two challenges, key to re-ID is to learn discriminative features. We argue that such features need to be of \emph{omni-scale}, defined as the combination of variable homogeneous scales and heterogeneous scales, each of which is composed of a mixture of multiple scales. The need for omni-scale features is evident from Fig.~\ref{fig:examples}. To match people and distinguish them from impostors, features corresponding small local regions (e.g.~shoes, glasses) and global whole body regions are equally important. For example, given the query image in Fig.~\ref{fig:examples}(a) (left), looking at the global-scale features (e.g.~young man, a white T-shirt + grey shorts combo) would narrow down the search to the true match (middle) and an impostor (right). Now the local-scale features come into play. The shoe region gives away the fact that the person on the right is an impostor (trainers vs.~sandals). However, for more challenging cases, even features of variable homogeneous scales would not be enough and more complicated and richer features that span multiple scales are required. For instance, to eliminate the impostor in Fig.~\ref{fig:examples}(d) (right), one needs features that represent a white T-shirt with a specific logo in the front. Note that the logo is not distinctive on its own -- without the white T-shirt as context, it can be confused with many other patterns. Similarly, the white T-shirt is likely everywhere in summer (e.g.~Fig.~\ref{fig:examples}(a)). It is however the unique combination, captured by heterogeneous-scale features spanning both small (logo size) and medium (upper body size) scales, that makes the features most effective.

Nevertheless, none of the existing re-ID models addresses omni-scale feature learning. In recent years, deep convolutional neural networks (CNNs) have been widely used in person re-ID to learn discriminative features~\cite{chang2018multi,li2018harmonious,liu2017hydraplus,si2018dual,sun2018beyond,xu2018attention,yang2019towards,zheng2019joint}. However, most of the CNNs adopted, such as ResNet~\cite{he2016deep}, were originally designed for object category-level recognition tasks that are fundamentally different from the instance-level recognition task in re-ID. For the latter, omni-scale features are more important, as explained earlier. A few attempts at learning multi-scale features also exist~\cite{qian2017multi,chang2018multi}. Yet, none has the ability to learn features of both homogeneous and heterogeneous scales.

\begin{figure}[t]
\centering
\includegraphics[width=\columnwidth]{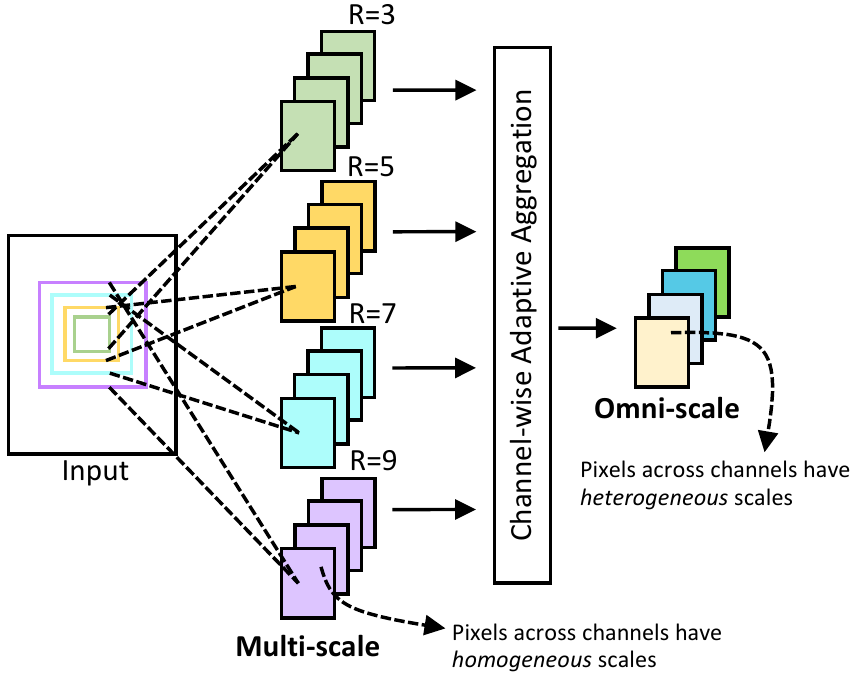}
\caption{A schematic of the proposed building block for OSNet. R: Receptive field size.}
\vspace{-0.5cm}
\label{fig:motivation}
\end{figure}

In this paper, we present \emph{OSNet}, a novel CNN architecture designed for learning omni-scale feature representations.
The underpinning building block consists of multiple convolutional streams with different receptive field sizes\footnote{We use scale and receptive field interchangeably.} (see Fig.~\ref{fig:motivation}). The feature scale that each stream focuses on is determined by \emph{exponent}, a new dimension factor that is linearly increased across streams to ensure that various scales are captured in each block. Critically, the resulting multi-scale feature maps are dynamically fused by channel-wise weights that are generated by a unified aggregation gate (AG).
The AG is a mini-network sharing parameters across all streams with a number of desirable properties for effective model training. With the trainable AG, the generated channel-wise weights become input-dependent, hence the dynamic scale fusion.
This novel AG design allows the network to learn omni-scale feature representations: depending on the specific input image, the gate could focus on a single scale by assigning a dominant weight to a particular stream or scale; alternatively, it can pick and mix and thus produce heterogeneous scales.

Apart from omni-scale feature learning, another key design principle adopted in OSNet is to construct a \emph{lightweight} network. This brings a couple of benefits:
(1) re-ID datasets are often of moderate size due to the difficulties in collecting across-camera matched person images. A lightweight network with a small number of parameters is thus less prone to overfitting.
(2) In a large-scale surveillance application (e.g.~city-wide surveillance using thousands of cameras), the most practical way for re-ID is to perform feature extraction at the camera end. Instead of sending the raw videos to a central server, only the extracted features need to be sent. For on-device processing, small re-ID networks are clearly preferred. To this end, in our building block, we factorise standard convolutions with pointwise and depthwise convolutions~\cite{howard2017mobilenets,sandler2018mobilenetv2}. 
The \textbf{contributions} of this work are thus both \textit{the concept of omni-scale feature learning} and \textit{an effective and efficient implementation of it in OSNet}.
The end result is a lightweight re-ID model that is more than one order of magnitude smaller than the popular ResNet50-based models, but performs better: OSNet achieves state-of-the-art performance on six person re-ID datasets, beating much larger networks, often by a clear margin. We also demonstrate the effectiveness of OSNet on object category recognition tasks, namely CIFAR~\cite{krizhevsky2009learning} and ImageNet~\cite{deng2009imagenet}, and a multi-label person attribute recognition task. The results suggest that omni-scale feature learning is useful beyond instance recognition and can be considered for a broad range of visual recognition tasks. Code and pre-trained models are available in Torchreid~\cite{torchreid}\footnote{\url{https://github.com/KaiyangZhou/deep-person-reid}}.

%%%%%%%%%%%%%%%%%%%%%%%%%%%
%%% Related work
%%%%%%%%%%%%%%%%%%%%%%%%%%%
\vspace{-0.2cm}
\section{Related Work} \label{sec:relatedwork}
\vspace{-0.2cm}

\keypoint{Deep re-ID architectures}.
Most existing deep re-ID CNNs~\cite{li2014deepreid,ahmed2015improved,varior2016gated,shen2018end,guo2018efficient,subramaniam2016deep,wang2018person} borrow architectures designed for generic object categorisation problems, such as ImageNet 1K object classification. Recently, some architectural modifications are introduced to reflect the fact that images in re-ID datasets contain instances of only one object category (i.e., person) that mostly stand upright. To exploit the upright body pose, \cite{sun2018beyond,zhang2017alignedreid,fu2019horizontal,wang2018learning} add auxiliary supervision signals to features pooled horizontally from the last convolutional feature maps. \cite{si2018dual,song2018mask,li2018harmonious} devise attention mechanisms to focus feature learning on the foreground person regions. In \cite{zhao2017spindle,su2017pose,xu2018attention,suh2018part,tian2018eliminating,zhang2019densely}, body part-specific CNNs are learned by means of off-the-shelf pose detectors. In \cite{li2017person,li2017learning,zhao2017deeply}, CNNs are branched to learn representations of global and local image regions. In \cite{yu2017devil,chang2018multi,liu2017hydraplus,wang2018resource}, multi-level features extracted at different layers are combined. However, \emph{none} of these re-ID networks learns multi-scale features explicitly at each layer of the networks as in our OSNet -- they typically rely on an external pose model and/or hand-pick specific layers for multi-scale learning. Moreover, heterogeneous-scale features computed from a mixture of different scales are not considered.

\keypoint{Multi-scale feature learning}.
As far as we know, the concept of omni-scale deep feature learning has never been introduced before. Nonetheless, the importance of multi-scale feature learning has been recognised recently and the multi-stream building block design has also been adopted. Compared to a number of re-ID networks with multi-stream building blocks~\cite{chang2018multi,qian2017multi}, OSNet is significantly different.
Specifically the layer design in \cite{chang2018multi} is based on ResNeXt~\cite{xie2017aggregated}, where each stream learns features at the same scale, while our streams in each block have different scales. Different to \cite{chang2018multi}, the network in \cite{qian2017multi} is built on Inception~\cite{szegedy2015going,szegedy2016rethinking}, where multiple streams were originally designed for low computational cost with handcrafted mixture of convolution and pooling layers. In contrast, our building block uses a scale-controlling factor to diversify the spatial scales to be captured.
Moreover, \cite{qian2017multi} fuses multi-stream features with learnable but fixed-once-learned streamwise weights only at the final block. Whereas we fuse multi-scale features within each building block using dynamic (input-dependent) channel-wise weights to learn combinations of multi-scale patterns. Therefore, only our OSNet is capable of learning omni-scale features with each feature channel potentially capturing discriminative features of either a single scale or a weighted mixture of multiple scales. Our experiments (see Sec.~\ref{subsec:evalOnReID}) show that OSNet significantly outperforms the models in \cite{chang2018multi,qian2017multi}.

\keypoint{Lightweight network designs}.
With embedded AI becoming topical, lightweight CNN design has attracted increasing attention. SqueezeNet~\cite{iandola2016squeezenet} compresses feature dimensions using $1 \times 1$ convolutions. IGCNet~\cite{zhang2017interleaved}, ResNeXt~\cite{xie2017aggregated} and CondenseNet~\cite{huang2018condense} leverage group convolutions. Xception~\cite{chollet2017xception} and MobileNet series~\cite{howard2017mobilenets,sandler2018mobilenetv2} are based on depthwise separable convolutions. Dense $1 \times 1$ convolutions are grouped with channel shuffling in ShuffleNet~\cite{zhang2018shufflenet}. In terms of lightweight design, our OSNet is similar to MobileNet by employing factorised convolutions, with some modifications that empirically work better for omni-scale feature learning.

%%%%%%%%%%%%%%%%%%%%%%%%%%%
%%% Proposed Approach
%%%%%%%%%%%%%%%%%%%%%%%%%%%
\vspace{-0.2cm}
\section{Omni-Scale Feature Learning} \label{sec:method}
In this section, we present OSNet, which specialises in learning omni-scale feature representations for the person re-ID task. We start with the factorised convolutional layer and then introduce the omni-scale residual block and the unified aggregation gate.

\subsection{Depthwise Separable Convolutions}
\begin{figure}[t]
\centering
\includegraphics[width=0.6\columnwidth]{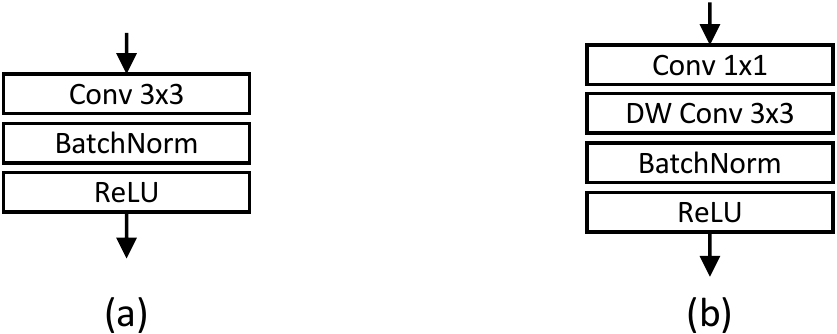}
\caption{(a) Standard $3\times3$ convolution. (b) \textit{Lite $3\times3$} convolution. DW: Depth-Wise.}
\label{fig:liteconv}
\vspace{-0.2cm}
\end{figure}

To reduce the number of parameters, we adopt the depthwise separable convolutions~\cite{howard2017mobilenets,chollet2017xception}. The basic idea is to divide a convolution layer $\mathrm{ReLU} (\bm{w} * \bm{x})$ with kernel $\bm{w} \in \mathbb{R}^{k \times k \times c \times c^\prime}$ into two separate layers $\mathrm{ReLU} ((\bm{v} \circ \bm{u}) * \bm{x})$ with depthwise kernel $\bm{u} \in \mathbb{R}^{k \times k \times 1 \times c^\prime}$ and pointwise kernel $\bm{v} \in \mathbb{R}^{1 \times 1 \times c \times c^\prime}$, where $*$ denotes convolution, $k$ the kernel size, $c$ the input channel width and $c^\prime$ the output channel width. Given an input tensor $\bm{x} \in \mathbb{R}^{h \times w \times c}$ of height $h$ and width $w$, the computational cost is reduced from $h \cdot w \cdot k^2 \cdot c \cdot c^\prime$ to $h \cdot w \cdot (k^2 + c) \cdot c^\prime$, and the number of parameters from $k^2 \cdot c \cdot c^\prime$ to $(k^2 + c) \cdot c^\prime$. In our implementation, we use $\mathrm{ReLU} ((\bm{u} \circ \bm{v}) * \bm{x})$ (pointwise $\rightarrow$ depthwise instead of depthwise $\rightarrow$ pointwise), which turns out to be more effective for omni-scale feature learning\footnote{The subtle difference between the two orders is when the channel width is increased: pointwise $\rightarrow$ depthwise increases the channel width before spatial aggregation.}. We call such layer \textit{Lite $3 \times 3$} hereafter. The implementation is shown in Fig.~\ref{fig:liteconv}.

\subsection{Omni-Scale Residual Block}
\begin{figure}[t]
\centering
\includegraphics[width=\columnwidth]{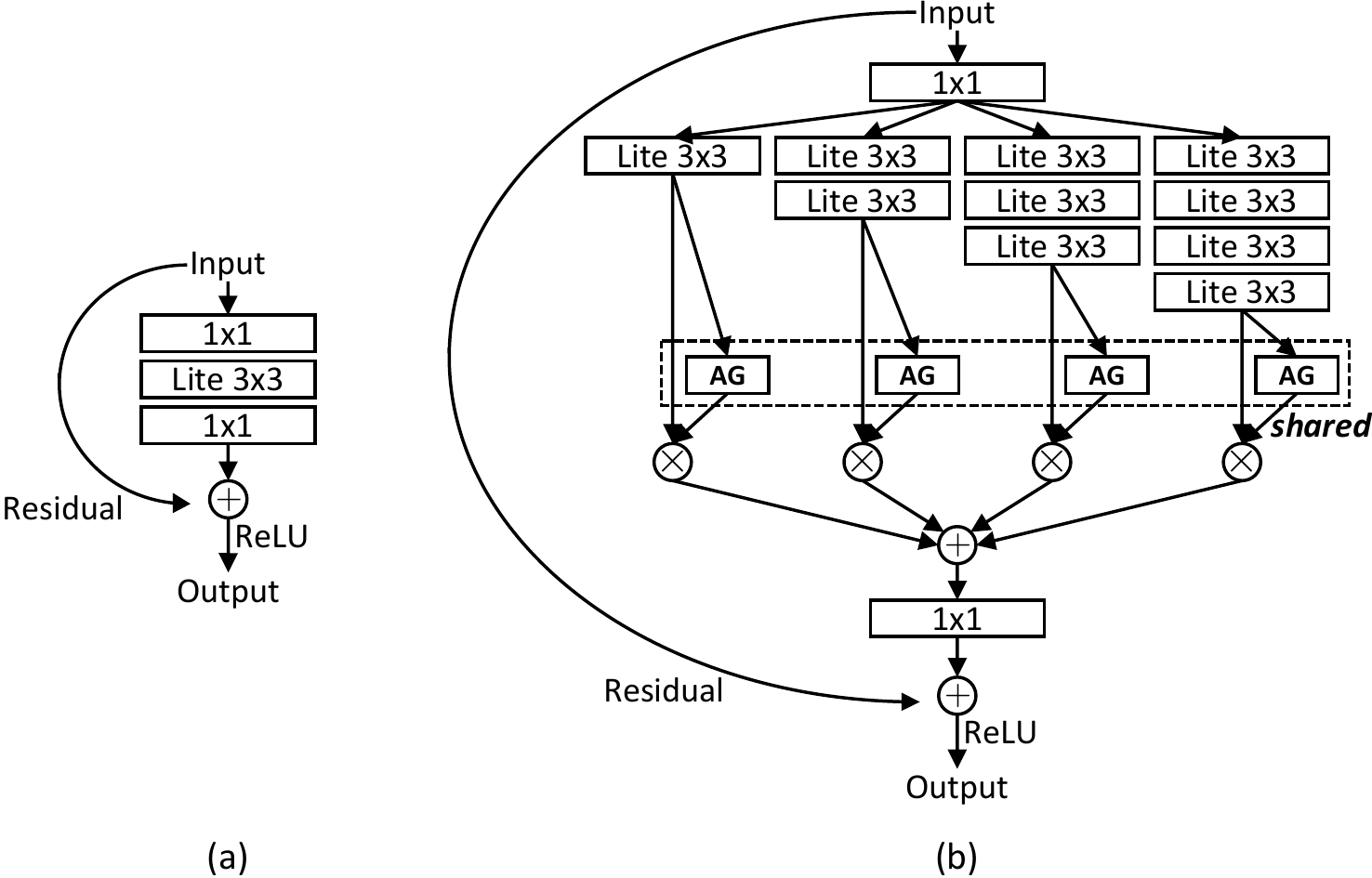}
\caption{
(a) Baseline bottleneck. (b) Proposed bottleneck. AG: Aggregation Gate. The first/last $1\times1$ layers are used to reduce/restore feature dimension.
}
\label{fig:bottleneck}
\vspace{-0.2cm}
\end{figure}

The building block in our architecture is the residual bottleneck~\cite{he2016deep}, equipped with the Lite $3\times3$ layer (see Fig.~\ref{fig:bottleneck}(a)). Given an input $\bm{x}$, this bottleneck aims to learn a residual $\tilde{\bm{x}}$ with a mapping function $F$, i.e.
\begin{equation} \label{eq:baseResidual}
\bm{y} = \bm{x} + \tilde{\bm{x}}, \quad \text{s.t.} \quad \tilde{\bm{x}} = F (\bm{x}), 
\end{equation}
where $F$ represents a Lite $3\times3$ layer that learns single-scale features (scale = 3). Note that here the  $1\times1$ layers are ignored in notation as they are used to manipulate feature dimension and do not contribute to the aggregation of spatial information~\cite{he2016deep,xie2017aggregated}.

\keypoint{Multi-scale feature learning}.
To achieve multi-scale feature learning, we extend the residual function $F$ by introducing a new dimension, {\it exponent} $t$, which represents the scale of the feature. For $F^t$, with $t > 1$, we stack $t$ Lite $3\times3$ layers, and this results in a receptive field of size $(2t+1)\times (2t+1)$. Then, the residual to be learned, $\tilde{\bm{x}}$, is the sum of incremental scales of representations up to $T$:
\begin{equation} \label{eq:mult_stream_residual}
\tilde{\bm{x}} = \sum_{t=1}^T F^t (\bm{x}), \quad \text{s.t.} \quad T \geqslant 1.
\end{equation}

When $T = 1$, Eq.~\ref{eq:mult_stream_residual} reduces to Eq.~\ref{eq:baseResidual} (see Fig.~\ref{fig:bottleneck}(a)). In this paper, our bottleneck is set with $T = 4$ (i.e. the largest receptive field is $9\times 9$) as shown in Fig.~\ref{fig:bottleneck}(b). The shortcut connection allows features at smaller scales learned in the current layer to be preserved effectively in the next layers, thus enabling the final features to capture a whole range of spatial scales.

\keypoint{Unified aggregation gate}.
So far, each stream can give us features of a specific scale, i.e., they are scale homogeneous. To learn omni-scale features, we propose to combine the outputs of different streams  in a dynamic way, i.e., different weights are assigned to different scales according to the input image, rather than being fixed after training.  
More specifically, the dynamic scale-fusion is achieved by a novel aggregation gate (AG), which is a \emph{learnable neural network}.

Let $\bm{x}^t$ denote $F^t (\bm{x})$, the omni-scale residual $\tilde{\bm{x}}$ is obtained by
\begin{equation} \label{eq:ag}
\tilde{\bm{x}} = \sum_{t=1}^T G (\bm{x}^t) \odot \bm{x}^t, \quad \text{s.t.} \quad \bm{x}^t \triangleq F^t (\bm{x}),
\end{equation}
where $G (\bm{x}^t)$ is a vector with length spanning the entire channel dimension of $\bm{x}^t$ and $\odot$ denotes the Hadamard product.
$G$ is implemented as a mini-network composed of a non-parametric global average pooling layer~\cite{lin2013network} and a multi-layer perceptron (MLP) with one ReLU-activated hidden layer, followed by the sigmoid activation.
To reduce parameter overhead, we follow~\cite{woo2018cbam,hu2018senet} to reduce the hidden dimension of the MLP with a reduction ratio, which is set to 16.

It is worth pointing out that, in contrast to using a single scalar-output function that provides a coarse scale-fusion, we choose to use channel-wise weights, i.e., the output of the AG network $G (\bm{x}^t)$ is a vector rather a scalar for the $t$-th stream. This design results in a more fine-grained fusion that tunes each feature channel. In addition, the weights are dynamically computed by being conditioned on the input data.  This is crucial for re-ID as the test images contain people of different identities from those in training; thus an adaptive/input-dependent feature-scale fusion strategy is more desirable.

Note that in our architecture, the AG is \emph{shared} for all feature streams in the same omni-scale residual block (dashed box in Fig.~\ref{fig:bottleneck}(b)). This is similar in spirit to the convolution filter parameter sharing in CNNs, resulting in a number of advantages. First, the number of parameters is independent of $T$ (number of streams), thus the model becomes more scalable. Second, unifying AG (sharing the same AG module across streams) has a nice property while performing backpropagation. Concretely, suppose the network is supervised by a loss function $\mathcal{L}$ which is differentiable and the gradient $\frac{\partial \mathcal{L}}{\partial \tilde{\bm{x}}}$ can be computed; the gradient w.r.t $G$, based on Eq.~\ref{eq:ag}, is
\begin{equation} \label{eq:gradChngate}
\frac{\partial \mathcal{L}}{\partial G} = \frac{\partial \mathcal{L}}{\partial \tilde{\bm{x}}} \frac{\partial \tilde{\bm{x}}}{\partial G}
= \frac{\partial \mathcal{L}}{\partial \tilde{\bm{x}}} ( \sum_{t=1}^T \bm{x}^t ).
\end{equation}

The second term in Eq.~\ref{eq:gradChngate} indicates that the supervision signals from all streams are gathered together to guide the learning of $G$. This desirable property disappears when each stream has its own gate.

\subsection{Network Architecture}
OSNet is constructed by simply stacking the proposed lightweight bottleneck layer-by-layer without any effort to customise the blocks at different depths (stages) of the network. The detailed network architecture is shown in Table~\ref{table:netarch}. For comparison, the same network architecture with standard convolutions has 6.9 million parameters and 3,384.9 million mult-add operations, which are $3 \times$ larger than our OSNet with the Lite $3\times3$ convolution layer design. The standard OSNet in Table~\ref{table:netarch} can be easily scaled up or down in practice, to balance model size, computational cost and performance. To this end,  we use a width multiplier\footnote{Width multiplier with magnitude smaller than 1 works on all layers in OSNet except the last FC layer whose feature dimension is fixed to 512.} and an image resolution multiplier, following \cite{howard2017mobilenets,sandler2018mobilenetv2,zhang2018shufflenet}.

\begin{table}[t]
\setlength{\tabcolsep}{8.5pt}
\centering
\small
\begin{tabular}{c|c|c}
\hline
stage & output & OSNet \\ \hline
\multirow{2}{*}{conv1} & 128$\times$64, 64 & 7$\times$7 conv, stride 2 \\
 & 64$\times$32, 64 & 3$\times$3 max pool, stride 2 \\ \hline
conv2 & 64$\times$32, 256 & bottleneck $\times$ 2 \\ \hline
\multirow{2}{*}{transition} & 64$\times$32, 256 & 1$\times$1 conv \\
 & 32$\times$16, 256 & 2$\times$2 average pool, stride 2 \\ \hline
conv3 & 32$\times$16, 384 & bottleneck $\times$ 2 \\ \hline
\multirow{2}{*}{transition} & 32$\times$16, 384 & 1$\times$1 conv \\ 
 & 16$\times$8, 384 & 2$\times$2 average pool, stride 2 \\ \hline
conv4 & 16$\times$8, 512 & bottleneck $\times$ 2 \\ \hline
conv5 & 16$\times$8, 512 & 1$\times$1 conv \\ \hline
gap & 1$\times$1, 512 & global average pool \\ \hline
fc & 1$\times$1, 512 & fc \\ \hline
\multicolumn{2}{c|}{\# params} & 2.2M \\ \hline
\multicolumn{2}{c|}{Mult-Adds} & 978.9M \\ \hline
\end{tabular}
\caption{Architecture of OSNet with input image size $256 \times 128$.}
\label{table:netarch}
\vspace{-0.2cm}
\end{table}

\keypoint{Relation to prior architectures}.
In terms of multi-stream design, OSNet is related to Inception~\cite{szegedy2015going} and ResNeXt~\cite{xie2017aggregated}, but has crucial differences  in several aspects.
First, the multi-stream design in OSNet strictly follows the scale-incremental principle dictated by the exponent (Eq.~\ref{eq:mult_stream_residual}). Specifically, different streams have different receptive fields but are built with the same Lite $3\times3$ layers (Fig.~\ref{fig:bottleneck}(b)). Such a design is more effective at capturing a wide range of scales. In contrast, Inception was originally designed to have low computational costs by sharing computations with multiple streams. Therefore its structure, which includes mixed operations of convolution and pooling, was handcrafted. ResNeXt has multiple equal-scale streams thus learning representations at the same scale.
Second, Inception/ResNeXt aggregates features by concatenation/addition while OSNet uses a unified AG (Eq.~\ref{eq:ag}), which facilitates the learning of combinations of multi-scale features. Critically, it means that the fusion is dynamic and adaptive to each individual input image. Therefore, OSNet's architecture is fundamentally different from that of Inception/ResNeXt in nature.
Third, OSNet uses factorised convolutions and thus the building block and subsequently the whole network is lightweight.
Compared with SENet~\cite{hu2018senet}, OSNet is conceptually different. Concretely, SENet aims to re-calibrate the feature channels by re-scaling the activation values for a single stream, whereas OSNet is designed to selectively fuse multiple feature streams of different receptive field sizes in order to learn omni-scale features (see Fig.~\ref{fig:motivation}).

%%%%%%%%%%%%%%%%%%%%%%%%%%%
%%% Experiments
%%%%%%%%%%%%%%%%%%%%%%%%%%%
\section{Experiments} \label{sec:exp}

\subsection{Evaluation on Person Re-Identification} \label{subsec:evalOnReID}
\vspace{-0.2cm}

\keypoint{Datasets and settings}.
We conduct experiments on six widely used person re-ID datasets: Market1501~\cite{zheng2015scalable}, CUHK03~\cite{li2014deepreid}, DukeMTMC-reID (Duke)~\cite{ristani2016performance,zheng2017unlabeled}, MSMT17~\cite{wei2018person}, VIPeR~\cite{gray2007evaluating} and GRID~\cite{loy2009multi}. Detailed dataset statistics are provided in Table~\ref{table:datasets}. The first four are considered as `big' datasets even though their sizes (around 30K training images for the largest MSMT17) are fairly moderate; while VIPeR and GRID are generally too small to train without using those big datasets for pre-training. For CUHK03, we use the 767/700 split~\cite{zhong2017rerank} with the detected images. For VIPeR and GRID, we first train a single OSNet from scratch using training images from Market1501, CUHK03, Duke and MSMT17 (Mix4), and then perform fine-tuning. Following \cite{li2017person}, the results on VIPeR and GRID are averaged over 10 random splits. Such a fine-tuning strategy has been commonly adopted by other deep learning approaches~\cite{liu2017hydraplus,wei2017glad,zhao2017spindle,li2017person,zhao2017deeply}. Cumulative matching characteristics (CMC) Rank-1 accuracy and mAP are used as evaluation metrics.

\begin{table}[t]
\setlength{\tabcolsep}{8pt}
\centering
\small
\begin{tabular}{l|c|c}
\hline
\multicolumn{1}{c|}{Dataset} & \# IDs (T-Q-G) & \# images (T-Q-G) \\
\hline
Market1501 & 751-750-751 & 12936-3368-15913 \\
CUHK03 & 767-700-700 & 7365-1400-5332 \\
Duke & 702-702-1110 & 16522-2228-17661 \\
MSMT17 & 1041-3060-3060 & 30248-11659-82161 \\
VIPeR & 316-316-316 & 632-632-632 \\
GRID & 125-125-900 & 250-125-900 \\
\hline
\end{tabular}
\caption{Dataset statistics. T: Train. Q: Query. G: Gallery.}
\vspace{-0.5cm}
\label{table:datasets}
\end{table}

\begin{table*}[h]
\setlength{\tabcolsep}{7pt}
\centering
\small
\begin{tabular}{l|c|c|cc|cc|cc|cc}
\hline
\multicolumn{1}{c|}{\multirow{2}{*}{Method}} & \multirow{2}{*}{Publication} & \multirow{2}{*}{Backbone} & \multicolumn{2}{c|}{Market1501} & \multicolumn{2}{c|}{CUHK03} & \multicolumn{2}{c|}{Duke} & \multicolumn{2}{c}{MSMT17} \\ \cline{4-11}
                                        & & & R1 & mAP & R1 & mAP & R1 & mAP & R1 & mAP \\
\hline
ShuffleNet$^{\dag \ddag}$~\cite{zhang2018shufflenet} & CVPR'18 & ShuffleNet & 84.8 & 65.0 & 38.4 & 37.2 & 71.6 & 49.9 & 41.5 & 19.9 \\
MobileNetV2$^{\dag \ddag}$~\cite{sandler2018mobilenetv2} & CVPR'18 & MobileNetV2 & 87.0 & 69.5 & \blue{46.5} & \blue{46.0} & 75.2 & 55.8 & \blue{50.9} & \blue{27.0} \\
BraidNet$^\dag$~\cite{wang2018person}    & CVPR'18 & BraidNet & 83.7 & 69.5 & - & - & 76.4 & 59.5 & - & - \\
HAN$^\dag$~\cite{li2018harmonious}   & CVPR'18 & Inception & \blue{91.2} & \blue{75.7} & 41.7 & 38.6 & \blue{80.5} & \blue{63.8} & - & - \\
OSNet$^\dag$ (ours) & ICCV'19 & OSNet & \red{93.6} & \red{81.0} & \red{57.1} & \red{54.2} & \red{84.7} & \red{68.6} & \red{71.0} & \red{43.3} \\ % scratch
\hline
\hline
DaRe~\cite{wang2018resource}          & CVPR'18 & DenseNet & 89.0 & 76.0 & 63.3 & 59.0 & 80.2 & 64.5 & - & - \\
PNGAN~\cite{qian2018pose}               & ECCV'18 & ResNet & 89.4 & 72.6 & - & - & 73.6 & 53.2 & - & - \\
KPM~\cite{shen2018end}                   & CVPR'18 & ResNet & 90.1 & 75.3 & - & - & 80.3 & 63.2 & - & - \\
MLFN~\cite{chang2018multi}               & CVPR'18 & ResNeXt & 90.0 & 74.3 & 52.8 & 47.8 & 81.0 & 62.8 & - & - \\
FDGAN~\cite{ge2018fd}                  & NeurIPS'18 & ResNet & 90.5 & 77.7 & - & - & 80.0 & 64.5 & - & - \\
DuATM~\cite{si2018dual}                 & CVPR'18 & DenseNet & 91.4 & 76.6 & - & - & 81.8 & 64.6 & - & - \\
Bilinear~\cite{suh2018part}             & ECCV'18 & Inception & 91.7 & 79.6 & - & - & 84.4 & 69.3 & - & - \\
G2G~\cite{shen2018deep}                  & CVPR'18 & ResNet & 92.7 & 82.5 & - & - & 80.7 & 66.4 & - & - \\
DeepCRF~\cite{chen2018group}          & CVPR'18 & ResNet & 93.5 & 81.6 & - & - & 84.9 & 69.5 & - & - \\
PCB~\cite{sun2018beyond}              & ECCV'18 & ResNet & 93.8 & 81.6 & 63.7 & 57.5 & 83.3 & 69.2 & 68.2 & 40.4 \\
SGGNN~\cite{shen2018person}            & ECCV'18 & ResNet & 92.3 & 82.8 & - & - & 81.1 & 68.2 & - & - \\
Mancs~\cite{wang2018mancs}             & ECCV'18 & ResNet & 93.1 & 82.3 & 65.5 & 60.5 & 84.9 & 71.8 & - & - \\
AANet~\cite{tay2019aanet}             & CVPR'19 & ResNet & 93.9 & 83.4 & - & - & \blue{87.7} & \blue{74.3} & - & - \\
CAMA~\cite{yang2019towards}            & CVPR'19 & ResNet & \blue{94.7} & 84.5 & \blue{66.6} & \blue{64.2} & 85.8 & 72.9 & - & - \\
IANet~\cite{hou2019interaction}        & CVPR'19 & ResNet & 94.4 & 83.1 & - & - & 87.1 & 73.4 & 75.5 & 46.8 \\
DGNet~\cite{zheng2019joint}              & CVPR'19 & ResNet & \red{94.8} & \red{86.0} & - & - & 86.6 & \red{74.8} & \blue{77.2} & \blue{52.3} \\
OSNet (ours) & ICCV'19 & OSNet & \red{94.8} & \blue{84.9} & \red{72.3} & \red{67.8} & \red{88.6} & 73.5 & \red{78.7} & \red{52.9} \\ % imagenet
\hline
\end{tabular}
\caption{
Results (\%) on big re-ID datasets. It is clear that OSNet achieves state-of-the-art performance on all datasets, surpassing most published methods by a clear margin. It is noteworthy that \emph{OSNet has only 2.2 million parameters}, which are far less than the current best-performing ResNet-based methods. -: not available. $\dag$: model trained from scratch. $\ddag$: reproduced by us. (Best and second best results in \red{red} and \blue{blue} respectively)
}
\vspace{-0.3cm}
\label{table:mainResults}
\end{table*}

\keypoint{Implementation details}.
A classification layer (linear FC + softmax) is mounted on the top of OSNet.
Training follows the standard classification paradigm where each person identity is regarded as a unique class. Similar to \cite{li2018harmonious,chang2018multi}, cross entropy loss with label smoothing~\cite{szegedy2016rethinking} is used for supervision.
For fair comparison against existing models, we implement two versions of OSNet. One is trained from scratch and the other is fine-tuned from ImageNet pre-trained weights.
Person matching is based on the $\ell_2$ distance of 512-D feature vectors extracted from the last FC layer (see Table \ref{table:netarch}).
Batch size and weight decay are set to 64 and 5e-4 respectively.
For training from scratch, SGD is used to train the network for 350 epochs. The learning rate starts from 0.065 and is decayed by 0.1 at 150, 225 and 300 epochs. Data augmentation includes random flip, random crop and random patch\footnote{RandomPatch works by (1) constructing a patch pool that stores randomly extracted image patches and (2) pasting a random patch selected from the patch pool onto an input image at random position.}.
For fine-tuning, we train the network with AMSGrad~\cite{reddi2018on} and initial learning rate of 0.0015 for 150 epochs. The learning rate is decayed by 0.1 every 60 epochs. During the first 10 epochs, the ImageNet pre-trained base network is frozen and only the randomly initialised classifier is open for training.
Images are resized to $256 \times 128$. Data augmentation includes random flip and random erasing~\cite{zhong2017random}.
The code is based on Torchreid~\cite{torchreid}.

\keypoint{Results on big re-ID datasets}.
From Table~\ref{table:mainResults}, we have the following observations.
(1) OSNet achieves state-of-the-art performance on all datasets, outperforming most published methods by a clear margin. It is evident from Table~\ref{table:mainResults} that the performance on re-ID benchmarks, especially Market1501 and Duke, has been saturated lately. Therefore, the improvements obtained by OSNet are significant. Crucially, the improvements are achieved with \emph{much smaller model size} -- most existing state-of-the-art re-ID models employ a ResNet50 backbone, which has more than 24 million parameters (considering their extra customised modules), while our OSNet has only 2.2 million parameters. This verifies the effectiveness of omni-scale feature learning for re-ID achieved by an extremely compact network. As OSNet is orthogonal to some methods, such as the image generation based DGNet~\cite{zheng2019joint}, they can be potentially combined to further boost the re-ID performance.
(2) OSNet yields strong performance with or without ImageNet pre-training. Among the very few existing lightweight re-ID models that can be trained from scratch (HAN and BraidNet), OSNet exhibits huge advantages. At R1, OSNet beats HAN/BraidNet by 2.4\%/9.9\% on Market1501 and 4.2\%/8.3\% on Duke. The margins at mAP are even larger. In addition, general-purpose lightweight CNNs are also compared without ImageNet pre-training. Table~\ref{table:mainResults} shows that OSNet surpasses the popular MobileNetV2 and ShuffleNet by large margins on all datasets. Note that all three networks have similar model sizes. These results thus demonstrate the versatility of our OSNet: It enables effective feature tuning from generic object categorisation tasks and offers robustness against model over-fitting when trained from scratch on datasets of moderate sizes.
(3) Compared with re-ID models that deploy a multi-scale/multi-stream architecture, namely those with a Inception or ResNeXt backbone \cite{li2018harmonious,su2017pose,chen2017person,wei2017glad,chang2018multi,si2018dual}, OSNet is clearly superior. As analysed in Sec.~\ref{sec:method}, this is attributed to the unique ability of OSNet to learn heterogeneous-scale features by combining multiple homogeneous-scale features with the dynamic AG.

\keypoint{Results on small re-ID datasets}.
VIPeR and GRID are very challenging datasets for deep re-ID approaches because they have only hundreds of training images - training on the large re-ID datasets and fine-tuning on them is thus necessary.
Table~\ref{table:resOnSmallData} compares OSNet with six state-of-the-art deep re-ID methods.
On VIPeR, it can be observed that OSNet outperforms the alternatives by a significant margin -- more than 11.4\% at R1. GRID is much more challenging than VIPeR because it has only 125 training identities (250 images) and extra distractors. Further, it was captured by real (operational) analogue CCTV cameras installed in busy public spaces. JLML~\cite{li2017person} is currently the best published method on GRID. It is noted that OSNet is marginally better than JLML on GRID. Overall, the strong performance of OSNet on these two small datasets is indicative of its practical usefulness in real-world applications where collecting large-scale training data is unscalable.

\begin{table}[t]
\centering
\small
\begin{tabular}{l|c|c|c}
\hline
\multicolumn{1}{c|}{Method} & Backbone & VIPeR & GRID \\
\hline
MuDeep~\cite{qian2017multi} & Inception & 43.0 & - \\ % (ICCV'17)
DeepAlign~\cite{zhao2017deeply} & Inception & 48.7 & - \\ % (ICCV'17)
JLML~\cite{li2017person} & ResNet & 50.2 & 37.5 \\ % (IJCAI'17)
Spindle~\cite{zhao2017spindle} & Inception & 53.8 & - \\ % (CVPR'17)
GLAD~\cite{wei2017glad} & Inception & 54.8 & - \\ % (ACMMM'17)
HydraPlus-Net~\cite{liu2017hydraplus} & Inception & 56.6 & - \\ % (ICCV'17)
OSNet (ours) & OSNet & \textbf{68.0} & \textbf{38.2} \\
\hline
\end{tabular}
\caption{Comparison with deep learning approaches on VIPeR and GRID. Only Rank-1 accuracy (\%) is reported. -: not available.}
\vspace{-0.3cm}
\label{table:resOnSmallData}
\end{table}

\begin{table}[t]
\setlength{\tabcolsep}{4pt}
\centering
\small
\begin{tabular}{c|l|c|c}
\hline
\multirow{2}{*}{Model} & \multicolumn{1}{c|}{\multirow{2}{*}{Architecture}} & \multicolumn{2}{c}{Market1501 } \\ \cline{3-4}
 & & R1 & mAP \\
\hline
\ablStdModel \label{ablStdModel:4str+uag} & $T=4$ + unified AG (\textbf{primary model}) & 93.6 & 81.0 \\
\ablStdModel \label{ablStdModel:4str(fullconv)+uag} & $T=4$ w/ full conv + unified AG & 94.0 & 82.7\\
\ablStdModel \label{ablStdModel:4str(t=1)+uag} & $T=4$ (same depth) + unified AG & 91.7 & 77.9 \\
\ablStdModel \label{ablStdModel:4str+cat} & $T=4$ + concatenation & 91.4 & 77.4 \\
\ablStdModel \label{ablStdModel:4str+add} & $T=4$ + addition & 92.0 & 78.2 \\
\ablStdModel \label{ablStdModel:4str+ags} & $T=4$ + separate AGs & 92.9 & 80.2 \\
\ablStdModel \label{ablStdModel:4str+uag(streamwise)} & $T=4$ + unified AG (stream-wise) & 92.6 & 80.0 \\
\ablStdModel \label{ablStdModel:4str+fixednum} & $T=4$ + learned-and-fixed gates & 91.6 & 77.5 \\
\ablStdModel \label{ablStdModel:1str} & $T=1$ & 86.5 & 67.7 \\
\ablStdModel \label{ablStdModel:2str+uag} & $T=2$ + unified AG & 91.7 & 77.0 \\
\ablStdModel \label{ablStdModel:3str+uag} & $T=3$ + unified AG & 92.8 & 79.9 \\
\hline
\end{tabular}
\caption{Ablation study on architectural design choices.}
\vspace{-0.4cm}
\label{table:ablationOnModelDesign}
\end{table}

\keypoint{Ablation experiments}.
Table~\ref{table:ablationOnModelDesign} evaluates our architectural design choices where our primary model is model~\ref{ablStdModel:4str+uag}. $T$ is the stream cardinality in Eq.~\ref{eq:mult_stream_residual}.
(1) {\it vs.~standard convolutions:}
Factorising convolutions reduces the R1 marginally by 0.4\% (model~\ref{ablStdModel:4str(fullconv)+uag} vs.~\ref{ablStdModel:4str+uag}). This means our architecture design maintains the representational power even though the model size is reduced by more than $3 \times$.
(2) {\it vs.~ResNeXt-like design:}
OSNet is transformed into a ResNeXt-like architecture by making all streams homogeneous in depth while preserving the unified AG, which refers to model~\ref{ablStdModel:4str(t=1)+uag}.
We observe that this variant is clearly outperformed by the primary model, with 1.9\%/3.1\% difference in R1/mAP. This further validates the necessity of our omni-scale design.
(3) {\it Multi-scale fusion strategy:}
To justify our design of the unified AG, we conduct experiments by changing the way how features of different scales are aggregated. The baselines are concatenation (model~\ref{ablStdModel:4str+cat}) and addition (model~\ref{ablStdModel:4str+add}). The primary model is better than the two baselines by more than 1.6\%/2.8\% at R1/mAP. Nevertheless, models~\ref{ablStdModel:4str+cat} and \ref{ablStdModel:4str+add} are still much better than the single-scale architecture (model~\ref{ablStdModel:1str}).
(4) {\it Unified AG vs.~separate AGs:}
When separate AGs are learned for each feature stream, the model size is increased and the nice property in gradient computation (Eq.~\ref{eq:gradChngate}) is lost. Empirically, unifying AG improves by 0.7\%/0.8\% at R1/mAP (model~\ref{ablStdModel:4str+uag} vs.~\ref{ablStdModel:4str+ags}), despite having less parameters.
(5) {\it Channel-wise gates vs.~stream-wise gates:}
By turning the channel-wise gates into stream-wise gates (model \ref{ablStdModel:4str+uag(streamwise)}), both the R1 and the mAP decline by 1\%. As feature channels encapsulate sophisticated correlations and can represent numerous visual concepts~\cite{fong2018net2vec}, it is advantageous to use channel-specific weights.
(6) {\it Dynamic gates vs.~static gates:}
In model~\ref{ablStdModel:4str+fixednum}, feature streams are fused by static (learned-and-then-fixed) channel-wise gates to mimic the design in~\cite{qian2017multi}. As a result, the R1/mAP drops off by 2.0\%/3.5\% compared with that of dynamic gates (primary model).
Therefore, adapting the scale fusion for individual input images is essential.
(7) {\it Evaluation on stream cardinality:}
The results are substantially improved from $T=1$ (model~\ref{ablStdModel:1str}) to $T=2$ (model~\ref{ablStdModel:2str+uag}) and gradually progress to $T=4$ (model~\ref{ablStdModel:4str+uag}).

\keypoint{Model shrinking hyper-parameters}.
We can trade-off between model size, computations and performance by adjusting the width multiplier $\beta$ and the image resolution multiplier $\gamma$.
Table~\ref{table:varyWidthAndInputSize} shows that by keeping one multiplier fixed and shrinking the other, the R1 drops off {\it smoothly}.
It is worth noting that 92.2\% R1 accuracy is obtained by a much shrunken version of OSNet with \emph{merely 0.2M parameters and 82M mult-adds} ($\beta=0.25$).
Compared with the results in Table~\ref{table:mainResults}, we can see that the shrunken OSNet is still very competitive against the latest proposed models, most of which are $100 \times$ bigger in size. This indicates that OSNet has a great potential for efficient deployment in resource-constrained devices such as a surveillance camera with an AI processor.

\begin{table}[t]
\setlength{\tabcolsep}{7pt}
\centering
\small
\begin{tabular}{c|c|c|c|c|c}
\hline
\multirow{2}{*}{$\beta$} & \multirow{2}{*}{\# params} & \multirow{2}{*}{$\gamma$} & \multirow{2}{*}{Mult-Adds} & \multicolumn{2}{c}{Market1501 } \\ \cline{5-6}
 & & & & R1 & mAP \\
\hline
1.0 & 2.2M & 1.0 & 978.9M & 94.8 & 84.9 \\
0.75 & 1.3M & 1.0 & 571.8M & 94.5 & 84.1 \\
0.5 & 0.6M & 1.0 & 272.9M & 93.4 & 82.6 \\
0.25 & 0.2M & 1.0 & 82.3M & 92.2 & 77.8 \\
\hline
1.0 & 2.2M & 0.75 & 550.7M & 94.4 & 83.7 \\
1.0 & 2.2M & 0.5 & 244.9M & 92.0 & 80.3 \\
1.0 & 2.2M & 0.25 & 61.5M & 86.9 & 67.3 \\
\hline
0.75 & 1.3M & 0.75 & 321.7M & 94.3 & 82.4 \\
0.75 & 1.3M & 0.5 & 143.1M & 92.9 & 79.5 \\
0.75 & 1.3M & 0.25 & 35.9M & 85.4 & 65.5 \\
\hline
0.5 & 0.6M & 0.75 & 153.6M & 92.9 & 80.8 \\
0.5 & 0.6M & 0.5 & 68.3M & 91.7 & 78.5 \\
0.5 & 0.6M & 0.25 & 17.2M & 85.4 & 66.0 \\
\hline
0.25 & 0.2M & 0.75 & 46.3M & 91.6 & 76.1 \\
0.25 & 0.2M & 0.5 & 20.6M & 88.7 & 71.8  \\
0.25 & 0.2M & 0.25 & 5.2M & 79.1 & 56.0 \\
\hline
\end{tabular}
\caption{Results (\%) of varying width multiplier $\beta$ and resolution multiplier $\gamma$ for OSNet.
For input size, $\gamma=0.75$: $192\times96$; $\gamma=0.5$: $128\times64$; $\gamma=0.25$: $64\times32$.}
\vspace{-0.3cm}
\label{table:varyWidthAndInputSize}
\end{table}

\begin{figure*}[t]
\centering
\includegraphics[width=0.99\textwidth]{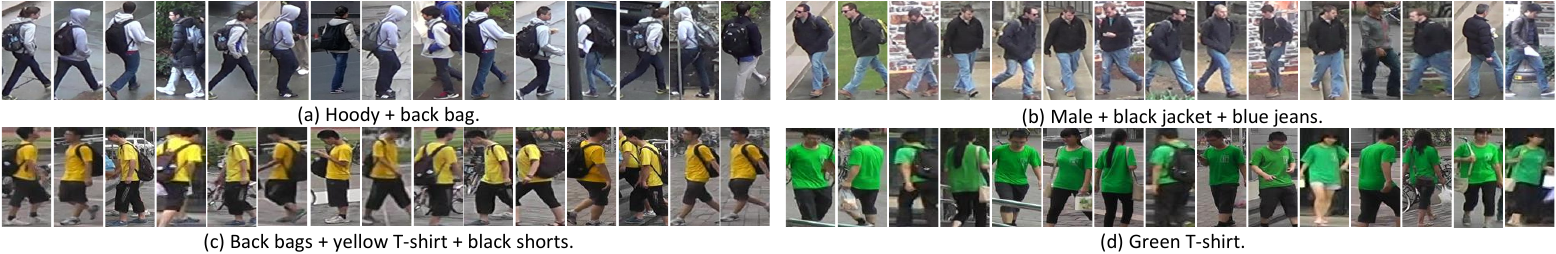}
\caption{Image clusters of similar gating vectors. The visualisation shows that our unified aggregation gate is capable of learning the combination of homogeneous and heterogeneous scales in a dynamic manner.}
\label{fig:clusterimgs}
\vspace{-0.5cm}
\end{figure*}

\keypoint{Visualisation of unified aggregation gate}.
As the gating vectors produced by the AG inherently encode the way how the omni-scale feature streams are aggregated, we can understand what the AG sub-network has learned by visualising images of similar gating vectors. To this end, we concatenate the gating vectors of four streams in the last bottleneck, perform k-means clustering on test images of Mix4, and select top-15 images closest to the cluster centres. Fig.~\ref{fig:clusterimgs} shows four example clusters where images within the same cluster exhibit similar patterns, i.e., combinations of global-scale and local-scale appearance.

\begin{figure}[t]
\centering
\includegraphics[width=0.9\columnwidth]{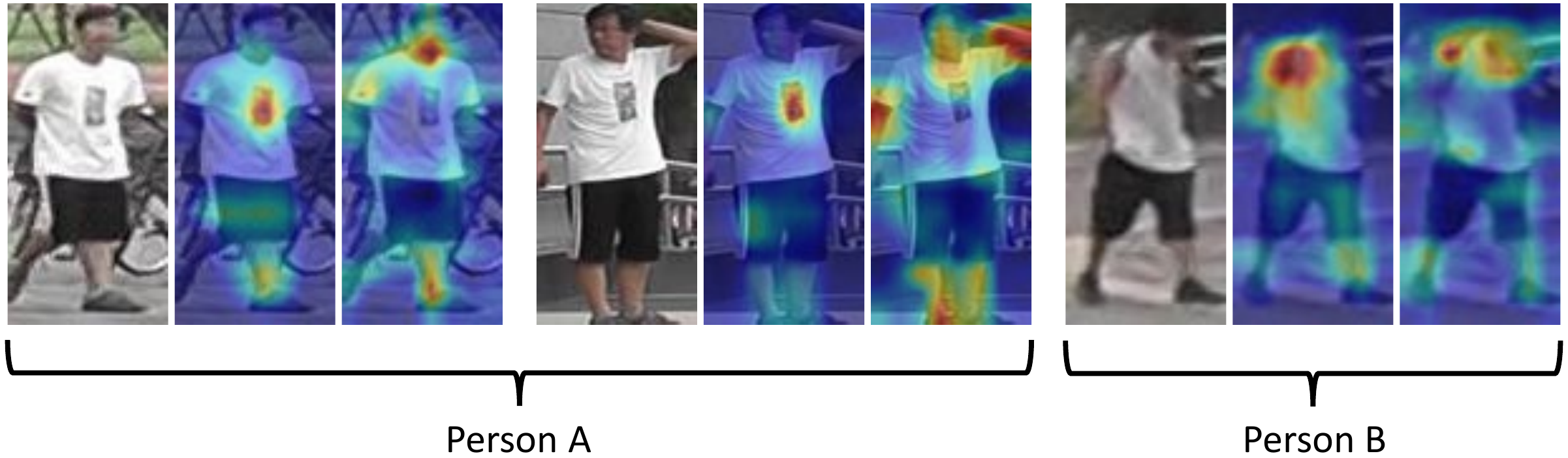}
\caption{Each triplet contains, from left to right, original image, activation map of OSNet and activation map of single-scale baseline. These images indicate that OSNet can detect subtle differences between visually similar persons.}
\label{fig:vis_actmap}
\vspace{-0.2cm}
\end{figure}

\keypoint{Visualisation of attention}.
To understand how our designs help OSNet learn discriminative features, we visualise the activations of the last convolutional feature maps to investigate where the network focuses on to extract features, i.e. attention.
Following \cite{zagoruyko2017paying}, the activation maps are computed as the sum of absolute-valued feature maps along the channel dimension followed by a spatial $\ell$2 normalisation.
Fig.~\ref{fig:vis_actmap} compares the activation maps of OSNet and the single-scale baseline (model~\ref{ablStdModel:1str} in Table~\ref{table:ablationOnModelDesign}). It is clear that OSNet can capture the local discriminative patterns of Person A (e.g., the clothing logo) which distinguish Person A from Person B. In contrast, the single-scale model over-concentrates on the face region, which is unreliable for re-ID due to the low resolution of surveillance images. Therefore, this qualitative result shows that our multi-scale design and unified aggregation gate enable OSNet to identify subtle differences between visually similar persons -- a vital requirement for accurate re-ID.

%%%%%%%%%%%%%%%%%%%%%%%%%%%%%%%%%%%%%%%
\subsection{Evaluation on Person Attribute Recognition}
Although person attribute recognition is a category-recognition problem, it is closely related to the person re-ID problem in that omni-scale feature learning is also critical: some attributes such as `view angle' are global; others such as `wearing glasses' are local; heterogeneous-scale features are also needed for recognising attributes such as `age'.

\keypoint{Datasets and settings}.
We use PA-100K~\cite{liu2017hydraplus}, the largest person attribute recognition dataset. PA-100K contains 80K training images and 10K test images. Each image is annotated with 26 attributes, e.g., male/female, wearing glasses, carrying hand bag. Following \cite{liu2017hydraplus}, we adopt five evaluation metrics, including mean Accuracy (mA), and four instance-based metrics, namely Accuracy (Acc), Precision (Prec), Recall (Rec) and F1-score (F1). Please refer to \cite{li2016richly} for the detailed definitions.

\keypoint{Implementation details}.
A sigmoid-activated attribute prediction layer is added on the top of OSNet. Following \cite{li2015multi,liu2017hydraplus}, we use the weighted multi-label classification loss for supervision. For data augmentation, we adopt random translation and mirroring. OSNet is trained from scratch with SGD, momentum of 0.9 and initial learning rate of 0.065 for 50 epochs. The learning rate is decayed by 0.1 at 30 and 40 epochs.

\begin{figure}[t]
\centering
\includegraphics[width=0.9\columnwidth]{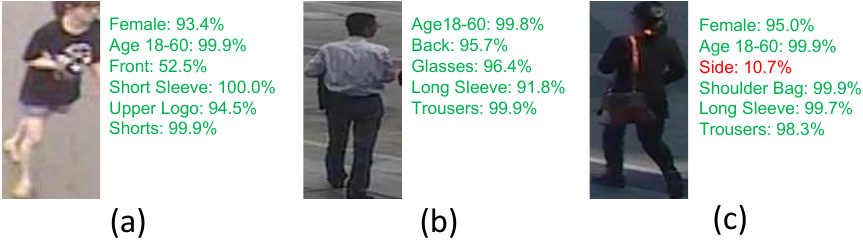}
\caption{Likelihoods on ground-truth attributes predicted by OSNet. Correct/incorrect classifications based on threshold 50\% are shown in \green{green}/\red{red}.}
\label{fig:attrImgs}
\vspace{-0.5cm}
\end{figure}

\keypoint{Results}.
Table \ref{table:resOnAttrRecog} compares OSNet with two state-of-the-art methods \cite{li2015multi,liu2017hydraplus} on PA-100K. It can be seen that OSNet outperforms both alternatives on all five evaluation metrics. 
Fig.~\ref{fig:attrImgs} provides some qualitative results. It shows that OSNet is particularly strong at predicting attributes that can only be inferred by examining features of heterogeneous scales such as age and gender.

\begin{table}[t]
\setlength{\tabcolsep}{5pt}
\centering
\small
\begin{tabular}{l|c|c|c|c|c}
\hline
\multicolumn{1}{c|}{Method} & \multicolumn{5}{c}{PA-100K} \\ \cline{2-6}
 & mA & Acc & Prec & Rec & F1 \\
\hline
DeepMar \cite{li2015multi} & 72.7 & 70.4 & 82.2 & 80.4 & 81.3 \\
HydraPlusNet \cite{liu2017hydraplus} & 74.2 & 72.2 & 83.0 & 82.1 & 82.5 \\
\hline
OSNet & \textbf{74.6} & \textbf{76.0} & \textbf{88.3} & \textbf{82.5} & \textbf{85.3} \\
\hline
\end{tabular}
\caption{Results (\%) on pedestrian attribute recognition.}
\vspace{-0.3cm}
\label{table:resOnAttrRecog}
\end{table}

%%%%%%%%%%%%%%%%%%%%%%%%%%%%%%%%%%%%%%%
\subsection{Evaluation on CIFAR} \label{subsec:evalOnImgClass}

\keypoint{Datasets and settings}.
CIFAR10/100~\cite{krizhevsky2009learning} has 50K training images and 10K test images, each with the size of $32\times32$. OSNet is trained following the setting in \cite{he2016identity,zagoruyko2016wide}.
Apart from the default OSNet in Table \ref{table:netarch}, a deeper version is constructed by increasing the number of staged bottlenecks from 2-2-2 to 3-8-6. Error rate is reported as the metric.

\keypoint{Results}.
Table~\ref{table:resOnCifar} compares OSNet with a number of state-of-the-art object recognition models. The results suggest that, although OSNet is originally designed for fine-grained object instance recognition task in re-ID, it is also highly competitive on object category recognition tasks. Note that CIFAR100 is more difficult than CIFAR10 because it contains ten times fewer training images per class (500 vs.~5,000). However, OSNet's performance on CIFAR100 is stronger, indicating that it is better at capturing useful patterns with limited data, hence its excellent performance on the data-scarce re-ID benchmarks.

\begin{table}[t]
\setlength{\tabcolsep}{3pt}
\centering
\small
\begin{tabular}{l|c|c|c|c}
\hline
\multicolumn{1}{c|}{Method} & Depth & \# params & CIFAR10 & CIFAR100 \\
\hline
pre-act ResNet~\cite{he2016identity} & 164 & 1.7M & 5.46 & 24.33 \\
pre-act ResNet~\cite{he2016identity} & 1001 & 10.2M & 4.92 & 22.71 \\
Wide ResNet~\cite{zagoruyko2016wide} & 40 & 8.9M & 4.97 & 22.89 \\ % k=4
Wide ResNet~\cite{zagoruyko2016wide} & 16 & 11.0M & 4.81 & 22.07 \\ % k=8
DenseNet~\cite{huang2017densely} & 40 & 1.0M & 5.24 & 24.42 \\
DenseNet~\cite{huang2017densely} & 100 & 7.0M & \textbf{4.10} & 20.20 \\
\hline
OSNet & 78 & 2.2M & 4.41 & 19.21 \\
OSNet & 210 & 4.6M & 4.18 & \textbf{18.88} \\
\hline
\end{tabular}
\caption{Error rates (\%) on CIFAR datasets. All methods here use translation and mirroring for data augmentation. Pointwise and depthwise convolutions are counted as separate layers.}
\label{table:resOnCifar}
\vspace{-0.3cm}
\end{table}

\keypoint{Ablation study}.
We compare our primary model with model~\ref{ablStdModel:1str} (single-scale baseline in Table~\ref{table:ablationOnModelDesign}) and model~\ref{ablStdModel:4str+add} (four streams + addition) on CIFAR10/100. Table~\ref{table:ablationOnCifar} shows that both omni-scale feature learning and unified AG contribute positively to the overall performance of OSNet.
\begin{table}[t]
\centering
\small
\begin{tabular}{l|c|c}
\hline
\multicolumn{1}{c|}{Architecture} & CIFAR10 & CIFAR100 \\
\hline
$T=1$ & 5.49 & 21.78 \\
$T=4$ + addition & 4.72 & 20.24 \\
$T=4$ + unified AG & \textbf{4.41} & \textbf{19.21} \\
\hline
\end{tabular}
\caption{Ablation study on OSNet on CIFAR10/100.}
\label{table:ablationOnCifar}
\vspace{-0.5cm}
\end{table}

\begin{table}[t]
\centering
\small
\begin{tabular}{l|c|c|c|c}
\hline
\multicolumn{1}{c|}{Method} & $\beta$ & \# params & Mult-Adds & Top1 \\
\hline
SqueezeNet~\cite{iandola2016squeezenet} & 1.0 & 1.2M & - & 57.5 \\
%\hline
MobileNetV1~\cite{howard2017mobilenets} & 0.5 & 1.3M & 149M & 63.7 \\
MobileNetV1~\cite{howard2017mobilenets} & 0.75 & 2.6M & 325M & 68.4 \\
MobileNetV1~\cite{howard2017mobilenets} & 1.0 & 4.2M & 569M & 70.6 \\
%\hline
ShuffleNet~\cite{zhang2018shufflenet} & 1.0 & 2.4M & 140M & 67.6 \\
ShuffleNet~\cite{zhang2018shufflenet} & 1.5 & 3.4M & 292M & 71.5 \\
ShuffleNet~\cite{zhang2018shufflenet} & 2.0 & 5.4M & 524M & 73.7 \\
%\hline
MobileNetV2~\cite{sandler2018mobilenetv2} & 1.0 & 3.4M & 300M & 72.0 \\
MobileNetV2~\cite{sandler2018mobilenetv2} & 1.4 & 6.9M & 585M & 74.7 \\
OSNet (ours) & 0.5 & 1.1M & 424M & 69.5 \\
OSNet (ours) & 0.75 & 1.8M & 885M & 73.5 \\
OSNet (ours) & 1.0 & 2.7M & 1511M & {\bf 75.5} \\
\hline
\end{tabular}
\caption{Single-crop top1 accuracy (\%) on ImageNet-2012 validation set. $\beta$: width multiplier. M: Million.}
\vspace{-0.3cm}
\label{table:resOnImagenet}
\end{table}

\subsection{Evaluation on ImageNet}
In this section, the results on the larger-scale ImageNet 1K category dataset (LSVRC-2012 \cite{deng2009imagenet}) are presented. 

\keypoint{Implementation}.
OSNet is trained with SGD, initial learning rate of 0.4, batch size of 1024 and weight decay of 4e-5 for 120 epochs. For data augmentation, we use random $224\times224$ crops on $256\times256$ images and random mirroring. To benchmark, we report single-crop\footnote{$224\times224$ centre crop from $256\times256$.} top1 accuracy on the LSVRC-2012 validation set~\cite{deng2009imagenet}.

\keypoint{Results}.
Table~\ref{table:resOnImagenet} shows that OSNet outperforms the alternative lightweight models by a clear margin.
In particular OSNet$\times$1.0 surpasses MobiltNetV2$\times$1.0 by 3.5\% and MobiltNetV2$\times$1.4 by 0.8\%. It is noteworthy that MobiltNetV2$\times$1.4 is around 2.5$\times$ larger than our OSNet$\times$1.0. OSNet$\times$0.75 performs on par with ShuffleNet$\times$2.0 and outperforms ShuffleNet$\times$1.5/$\times$1.0 by 2.0\%/5.9\%. These results give a strong indication that OSNet has a great potential for a broad range of visual recognition tasks.
Note that although the model size is smaller, our OSNet does have a higher number of mult-adds operations than its main competitors. This is mainly due to the multi-stream design. However, if both model size and number of Multi-Adds need to be small for a certain application, we can reduce the latter by introducing pointwise convolutions with group convolutions and channel shuffling~\cite{zhang2018shufflenet}.
The overall results on CIFAR and ImageNet show that omni-scale feature learning is beneficial beyond re-ID and should be considered for a broad range of visual recognition tasks.

%%%%%%%%%%%%%%%%%%%%%%%%%%%
%%% Conclusion
%%%%%%%%%%%%%%%%%%%%%%%%%%%
\vspace{-0.2cm}
\section{Conclusion}
\vspace{-0.2cm}
We presented OSNet, a lightweight CNN architecture that is capable of learning omni-scale feature representations.
Extensive experiments on six person re-ID datasets demonstrated that OSNet achieved state-of-the-art performance, despite its lightweight design.
The superior performance on object categorisation tasks and a multi-label attribute recognition task further suggested that OSNet is of wide interest to visual recognition beyond re-ID.

%% file: supp.tex
%%%%%%%%% BODY TEXT

%%%%%%%%%%%%%%%%%
% Supplementary
%%%%%%%%%%%%%%%%%
\setcounter{section}{0}
\renewcommand\thesection{\Alph{section}}

\section*{Supplementary}
The results in the main paper have been presented at ICCV'19. In this supplementary, we show additional results to further demonstrate the stength of OSNet.

\begin{table*}[t]
\centering
\small
\begin{tabular}{l | c | c c c c || c | c c c c}
\hline
\multirow{2}{*}{Method} & \multirow{2}{*}{Source} & \multicolumn{4}{c||}{Target: Duke} & \multirow{2}{*}{Source} & \multicolumn{4}{c}{Target: Market1501} \\
 & & R1 & R5 & R10 & mAP & & R1 & R5 & R10 & mAP \\
\hline
MMFA~\cite{lin2018multi} & Market1501 + Duke ($U$) & 45.3 & 59.8 & 66.3 & 24.7 & Duke + Market1501 ($U$) & 56.7 & 75.0 & 81.8 & 27.4 \\
SPGAN~\cite{deng2018image} & Market1501 + Duke ($U$) & 46.4 & 62.3 & 68.0 & 26.2 & Duke + Market1501 ($U$) & 57.7 & 75.8 & 82.4 & 26.7 \\
TJ-AIDL~\cite{wang2018transferable} & Market1501 + Duke ($U$) & 44.3 & 59.6 & 65.0 & 23.0 & Duke + Market1501 ($U$) & 58.2 & 74.8 & 81.1 & 26.5 \\
ATNet~\cite{liu2019adaptive} & Market1501 + Duke ($U$) & 45.1 & 59.5 & 64.2 & 24.9 & Duke + Market1501 ($U$) & 55.7 & 73.2 & 79.4 & 25.6 \\
CamStyle~\cite{zhong2019camstyle} & Market1501 + Duke ($U$) & 48.4 & 62.5 & 68.9 & 25.1 & Duke + Market1501 ($U$) & 58.8 & 78.2 & 84.3 & 27.4 \\
HHL~\cite{zhong2018gen} & Market1501 + Duke ($U$) & 46.9 & 61.0 & 66.7 & 27.2 & Duke + Market1501 ($U$) & 62.2 & 78.8 & 84.0 & 31.4 \\
ECN~\cite{zhong2019invariance} & Market1501 + Duke ($U$) & 63.3 & 75.8 & 80.4 & 40.4 & Duke + Market1501 ($U$) & 75.1 & 87.6 & 91.6 & 43.0 \\
\rowcolor{lightgray} OSNet-IBN (ours) & Market1501 & 48.5 & 62.3 & 67.4 & 26.7 & Duke & 57.7 & 73.7 & 80.0 & 26.1 \\
\hline \hline
MAR~\cite{yu2019unsupervised} & MSMT17+Duke ($U$) & 67.1 & 79.8 & - & 48.0 & MSMT17+Market1501 ($U$) & 67.7 & 81.9 & - & 40.0 \\
PAUL~\cite{yang2019patch} & MSMT17+Duke ($U$) & 72.0 & 82.7 & 86.0 & 53.2 & MSMT17+Market1501 ($U$) & 68.5 & 82.4 & 87.4 & 40.1 \\
\rowcolor{lightgray} OSNet-IBN (ours) & MSMT17 & 67.4 & 80.0 & 83.3 & 45.6 & MSMT17 & 66.5 & 81.5 & 86.8 & 37.2 \\
\hline
\end{tabular}
\caption{Cross-domain re-ID results. It is worth noting that OSNet-IBN (highlighted rows), without using any target data, can achieve competitive performance with state-of-the-art unsupervised domain adaptation re-ID methods. $U$: Unlabelled.}
\label{table:xdomainReID}
\end{table*}

\section{A Strong Backbone for Cross-Domain Re-ID}
In this section, we construct a strong backbone model for cross-domain re-ID based on OSNet.
Following~\cite{pan2018ibn}, we add instance normalisation (IN)~\cite{ulyanov2017improved} to the lower layers (\texttt{conv1, conv2}) in OSNet. Specifically, IN is inserted after the residual connection and before the ReLU function in a bottleneck. It has been shown in \cite{pan2018ibn} that IN can improve the generalisation performance on cross-domain semantic segmentation tasks. Here we apply the same idea to OSNet and show that we can build a strong backbone model for cross-domain re-ID. We call this new network OSNet-IBN.

\keypoint{Settings}.
Following the recent works~\cite{zhong2019invariance,zhong2019camstyle,liu2019adaptive,yu2019unsupervised,yang2019patch}, we choose Market1501 and Duke as the target datasets. The source dataset is either Market1501, Duke or MSMT17\footnote{Following~\cite{yu2019unsupervised,yang2019patch}, all 126,441 images of 4,101 identities in MSMT17 are used for training.}.
Models are trained on labelled source data and directly tested on target data.

\keypoint{Implementation details}.
Similar to the conventional setting, we use cross-entropy loss as the objective function.
We train OSNet-IBN with AMSGrad~\cite{reddi2018on}, batch size of 64, weight decay of 5e-4 and initial learning rate of 0.0015 for 150 epochs. The learning rate is decayed by 0.1 every 60 epochs.
During the first 10 epochs, only the randomly initialised classification layer is open for training while the ImageNet pre-trained base network is frozen.
All images are resized to $256 \times 128$.
Data augmentation includes random flip and color jittering. We observed that random erasing~\cite{zhong2017random} dramatically decreased the cross-domain results so we did not use it.

\keypoint{Results}.
Table~\ref{table:xdomainReID} compares OSNet-IBN with current state-of-the-art unsupervised domain adaptation (UDA) methods.
It is clear that OSNet-IBN achieves highly competitive performance or even better results than some UDA methods on the target datasets, despite \textit{only using source data for training}.
In particular, on Market1501$\rightarrow$Duke (at R1), OSNet-IBN beats all the UDA methods except ECN; on MSMT17$\rightarrow$Duke, OSNet-IBN performs on par with MAR; on MSMT17$\rightarrow$Market1501, OSNet-IBN obtains comparable results with MAR and PAUL.
These results demonstrate that our OSNet-IBN, with a minor modification, can be used as a strong backbone model for cross-domain re-ID\footnote{See~\cite{zhou2019learning} for an improved OSNet-IBN (called OSNet-AIN~\cite{zhou2019learning}) which achieves better cross-domain performance via neural architecture search.}.

\begin{table*}[t]
\setlength{\tabcolsep}{3.5pt}
\centering
\small
\subfloat[Pixel normalisation parameters]{
\label{tab:recipe_pixel_norm}
\centering
\begin{tabular}[b]{l | c c | c c}
\hline
\multirow{2}{*}{Mean \& std from} & \multicolumn{2}{c|}{Market1501} & \multicolumn{2}{c}{Duke} \\
 & R1 & mAP & R1 & mAP \\ \hline
ImageNet & 94.6\std{0.1} & 86.5\std{0.2} & 88.6\std{0.3} & 76.6\std{0.1} \\
Re-ID dataset & 94.4\std{0.0} & 86.3\std{0.1} & 88.5\std{0.1} & 76.5\std{0.2} \\
\hline
\end{tabular}
}
~
%%%%%%%%%%%%%%%%%%%%%%%%%%%%
\subfloat[Input size]{
\label{tab:recipe_input_size}
\centering
\begin{tabular}[b]{l | c c | c c}
\hline
\multirow{2}{*}{Input size} & \multicolumn{2}{c|}{Market1501} & \multicolumn{2}{c}{Duke} \\
 & R1 & mAP & R1 & mAP \\ \hline
256$\times$128 & 94.6\std{0.1} & 86.5\std{0.2} & 88.6\std{0.3} & 76.6\std{0.1} \\
320$\times160$ & 94.9\std{0.1} & 86.9\std{0.1} & 88.5\std{0.5} & 76.8\std{0.2} \\
\hline
\end{tabular}
}

%%%%%%%%%%%%%%%%%%%%%%%%%%%%
\subfloat[Regularisation with entropy maximisation: $\mathcal{L}_{\rm ID} - \lambda_e \mathcal{L}_{\rm Entropy}$]{
\label{tab:recipe_entmax}
\centering
\begin{tabular}[b]{l | c c | c c}
\hline
\multirow{2}{*}{$\lambda_e$} & \multicolumn{2}{c|}{Market1501} & \multicolumn{2}{c}{Duke} \\
 & R1 & mAP & R1 & mAP \\ \hline
0 & 94.6\std{0.1} & 86.5\std{0.2} & 88.6\std{0.3} & 76.6\std{0.1} \\
0.01 & 94.6\std{0.1} & 86.4\std{0.1} & 88.5\std{0.5} & 76.5\std{0.3} \\
0.05 & 94.5\std{0.1} & 86.5\std{0.2} & 88.7\std{0.2} & 76.6\std{0.0} \\
0.1 & 94.7\std{0.1} & 86.4\std{0.3} & 88.4\std{0.1} & 76.7\std{0.2} \\
0.5 & 94.7\std{0.1} & 86.6\std{0.2} & 88.3\std{0.2} & 76.7\std{0.2} \\
\hline
\end{tabular}
}
~
%%%%%%%%%%%%%%%%%%%%%%%%%%%%
\subfloat[Deep mutual learning and model ensemble]{
\label{tab:recipe_dml}
\centering
\begin{tabular}[b]{l | c c | c c}
\hline
\multirow{2}{*}{} & \multicolumn{2}{c|}{Market1501} & \multicolumn{2}{c}{Duke} \\
 & R1 & mAP & R1 & mAP \\ \hline
w/o DML & 94.6\std{0.1} & 86.5\std{0.2} & 88.6\std{0.3} & 76.6\std{0.1} \\
w/ DML model-1 & 94.7\std{0.1} & 87.2\std{0.0} & 88.4\std{0.4} & 77.3\std{0.2} \\
w/ DML model-2 & 94.8\std{0.1} & 87.3\std{0.0} & 88.5\std{0.8} & 77.3\std{0.2} \\
w/ DML model-1+2 & 94.9\std{0.1} & 87.8\std{0.0} & 88.7\std{0.6} & 78.0\std{0.2} \\
\hline
\end{tabular}
}

%%%%%%%%%%%%%%%%%%%%%%%%%%%%
\subfloat[Auxiliary loss with hard example-mining triplet loss: $\mathcal{L}_{\rm ID} + \lambda_t \mathcal{L}_{\rm Triplet}$]{
\label{tab:recipe_triplet}
\centering
\begin{tabular}[b]{l | c c | c c}
\hline
\multirow{2}{*}{$\lambda_t$} & \multicolumn{2}{c|}{Market1501} & \multicolumn{2}{c}{Duke} \\
 & R1 & mAP & R1 & mAP \\ \hline
0 & 94.6\std{0.1} & 86.5\std{0.2} & 88.6\std{0.3} & 76.6\std{0.1} \\
0.1 & 94.7\std{0.4} & 86.7\std{0.2} & 88.3\std{0.3} & 76.9\std{0.1} \\
0.5 & 95.5\std{0.1} & 87.2\std{0.0} & 88.6\std{0.2} & 77.3\std{0.1} \\
1.0 & 94.9\std{0.1} & 86.9\std{0.1} & 88.4\std{0.1} & 76.8\std{0.4} \\
\hline
\end{tabular}
}
~
%%%%%%%%%%%%%%%%%%%%%%%%%%%%
\subfloat[$\mathcal{L}_{\rm Triplet}$ + deep mutual learning + model ensemble]{
\centering
\label{tab:recipe_triplet_dml}
\begin{tabular}[b]{l | c c | c c}
\hline
\multirow{2}{*}{} & \multicolumn{2}{c|}{Market1501} & \multicolumn{2}{c}{Duke} \\
 & R1 & mAP & R1 & mAP \\ \hline
$\lambda_t=0$ w/o DML & 94.6\std{0.1} & 86.5\std{0.2} & 88.6\std{0.3} & 76.6\std{0.1} \\
$\lambda_t=0.5$ + DML model-1 & 95.7\std{0.1} & 88.1\std{0.2} & 89.1\std{0.2} & 78.4\std{0.0} \\
$\lambda_t=0.5$ + DML model-2 & 95.5\std{0.2} & 88.0\std{0.1} & 89.6\std{0.3} & 78.5\std{0.2} \\
$\lambda_t=0.5$ + DML model-1+2 & 95.7\std{0.1} & 88.7\std{0.1} & 89.0\std{0.0} & 79.5\std{0.1} \\
\hline
\end{tabular}
}
\caption{Investigation on various training methods for improving OSNet's performance. All experiments are run for 3 times with different random seeds. Note that the implementation follows~\cite{zhou2019learning}, which is slightly different from the conference version.}
\end{table*}

\section{Training Recipes for Practitioners}
We investigate some training methods in order to further improve OSNet's performance. We not only show the methods that work, but also discuss what do not work in our experiments.

\keypoint{Implementation}.
We train the baseline OSNet following~\cite{zhou2019learning}, where the main difference compared with the conference version is the use of cosine annealing strategy~\cite{cosineLR} to decay the learning rate. For image matching, we use cosine distance. To make sure the result is convincing, we run \emph{every} experiment with 3 different random seeds and report the mean and standard deviation. We choose Market1501 and Duke for benchmarking.

\keypoint{Dataset-specific normalisation parameters}.
Most re-ID papers used the ImageNet mean and standard deviation for pixel normalisation, without justifying whether using dataset-specific statistics is a better choice. Typically, images from re-ID datasets exhibit drastic differences compared with the natural images from ImageNet, e.g., the person images for re-ID are usually of poor quality and blurred. Therefore, using the statistics from re-ID dataset for pixel normalisation seems to make more sense. However, Table~\ref{tab:recipe_pixel_norm} shows that the difference in performance is subtle, suggesting that collecting dataset-specific statistics might be unnecessary. In practice, we do, however, encourage practitioners to try both ways for their own datasets.

\keypoint{Will larger input size help}?
Table~\ref{tab:recipe_input_size} shows that using larger input size improves the performance, but only marginally. This is because OSNet can learn omni-scale features, which are insensitive to the input size. Considering that using $320 \times 160$ increases the flops from 978.9M to 1,529.3M, we suggest using $256 \times 128$.

\keypoint{Entropy maximisation}.
As re-ID datasets are small-scale, we add a entropy maximisation term~\cite{pereyra2017regularizing} to further regularise the network (this term penalises confident predictions). The results are shown in Table~\ref{tab:recipe_entmax} where we observe that this new regularisation term, with various balancing weights, has little effect on the performance.

\keypoint{Deep mutual learning (DML)}.
Zhang et al.~\cite{Zhang2017DeepML} has shown that DML can achieve notable improvement for re-ID (when using MobileNet~\cite{howard2017mobilenets}). We apply DML to training OSNet and report the results in Table~\ref{tab:recipe_dml}. It is clear that DML improves the mAP. This indicates that features learned with DML are more discriminative. As DML trains two networks simultaneously, it is natural to try model ensemble with these two networks. The results (last row in Table~\ref{tab:recipe_dml}) show clear improvements on both rank-1 and mAP. Note that when doing ensemble, we concatenate the features rather than performing mean-pooling. The latter makes more sense for classification tasks but not for retrieval tasks where features are used.

\keypoint{Auxiliary loss}.
Several recent re-ID approaches~\cite{Quan_2019_ICCV,Dai_2019_ICCV,Subramaniam_2019_ICCV} adopt a multi-loss training strategy, e.g., using both cross-entropy loss and triplet loss~\cite{hermans2017defense}. We investigate such training strategy for OSNet where a balancing weight $\lambda_t$ is added to scale the triplet loss (see the caption of Table~\ref{tab:recipe_triplet}). Table~\ref{tab:recipe_triplet} shows that the triplet loss improves the performance when $\lambda_t$ is carefully tuned. In practice, we encourage practitioners to use the cross-entropy loss as the main objective and the triplet loss as an auxiliary loss with a balancing weight (which needs to be tuned).

\keypoint{Combination}.
We combine the effective training techniques, i.e. DML and auxiliary loss learning (with the triplet loss), and show the results in Table~\ref{tab:recipe_triplet_dml}. It can be observed that the improvement is larger than that of using either technique alone. The best performance is obtained by fusing the two DML-trained models.

Therefore, we suggest training OSNet with cross-entropy loss + triplet loss ($\lambda_t=0.5$ as the rule of thumb) + DML and testing with model ensemble.